\documentclass[letterpaper, 10 pt, conference]{ieeeconf}
\IEEEoverridecommandlockouts
\usepackage[utf8]{inputenc}
\usepackage[T1]{fontenc}
\usepackage[english]{babel}
\usepackage{booktabs}
\usepackage{tabularx}
\usepackage{multirow}
\usepackage{listings}
\usepackage{subcaption}
\newcommand{\kw}[1]{\textbf{\textbf{#1}}}

\usepackage{amsmath}
\usepackage{amssymb}
\usepackage{amsthm}

\newtheoremstyle{main}
{1em}                                                
{1em}                                                
{\itshape}                                           
{0pt}                                                
{\scshape}                                           
{}                                                   
{2pt}                                                
{\thmname{#1}\thmnumber{ #2} [{\thmnote{\itshape #3}}]:~} 

\theoremstyle{main}

\usepackage{graphicx}

\usepackage{csquotes}
\usepackage{algorithm}
\usepackage{algpseudocode}
\usepackage[
	maxbibnames=99,
	maxcitenames=2,
	natbib=true,
	style=numeric-comp,
	backend=biber,
	sorting=none,
	giveninits=true,
	url=false,
	doi=false,
	eprint=false,
	isbn=false,
]{biblatex}

\usepackage[pdfa,colorlinks,bookmarksopen,bookmarksnumbered,allcolors=black,urlcolor=blue]{hyperref}

\usepackage[nameinlink,capitalise]{cleveref}
\crefname{line}{line}{lines}
\crefname{figure}{Fig.}{Figs.}
\Crefname{figure}{Fig.}{Figs.}
\crefname{equation}{Eq.}{Eqs.}
\Crefname{equation}{Eq.}{Eqs.}
\crefname{section}{Sec.}{Secs.}
\Crefname{section}{Sec.}{Secs.}
\crefname{definition}{Def.}{Defs.}
\Crefname{definition}{Def.}{Defs.}
\crefname{algorithm}{Alg.}{Algs.}
\Crefname{algorithm}{Alg.}{Algs.}
\crefname{assumption}{Asm.}{Asms.}
\Crefname{assumption}{Asm.}{Asms.}
\crefname{subassumption}{Asm.}{Asms.}
\Crefname{subassumption}{Asm.}{Asms.}
\Crefname{problem}{Problem}{Problems}
\crefname{problem}{Problem}{Problems}

\usepackage{pbalance}

\title{\LARGE \bf
Differentiable Particle Optimization for Fast Sequential Manipulation
}

\author{
Lucas Chen, Shrutheesh R. Iyer, and Zachary Kingston%
\thanks{LC, SRI, and ZK are with Department of Computer Science, Purdue University, {\tt \{chen4007, iyer480, zkingston\}@purdue.edu}}%
\\
}

\begin{document}
\maketitle

\begin{abstract}
Sequential robot manipulation tasks require finding collision-free trajectories that satisfy geometric constraints across multiple object interactions in potentially high-dimensional configuration spaces.
Solving these problems in real-time and at large scales has remained out of reach due to computational requirements.
Recently, GPU-based acceleration has shown promising results, but prior methods achieve limited performance due to CPU-GPU data transfer overhead and complex logic that prevents full hardware utilization.
To this end, we present SPaSM (Sampling Particle optimization for Sequential Manipulation), a fully GPU-parallelized framework that compiles constraint evaluation, sampling, and gradient-based optimization into optimized CUDA kernels for end-to-end trajectory optimization without CPU coordination.
The method consists of a two-stage particle optimization strategy: first solving placement constraints through massively parallel sampling, then lifting solutions to full trajectory optimization in joint space.
Unlike hierarchical approaches, SPaSM jointly optimizes object placements and robot trajectories to handle scenarios where motion feasibility constrains placement options.
Experimental evaluation on challenging benchmarks demonstrates solution times in the realm of \textbf{milliseconds} with a 100\% success rate; a $4000\times$ speedup compared to existing approaches. Code and examples are available at \href{https://commalab.org/papers/spasm}{commalab.org/papers/spasm}. 
\end{abstract}

\section{Introduction}
\begin{figure*}[t]
\vspace{1em}
    \centering
    \includegraphics[clip, trim=0.0cm 1.0cm 0.0cm 1.1cm, width=0.99\linewidth]{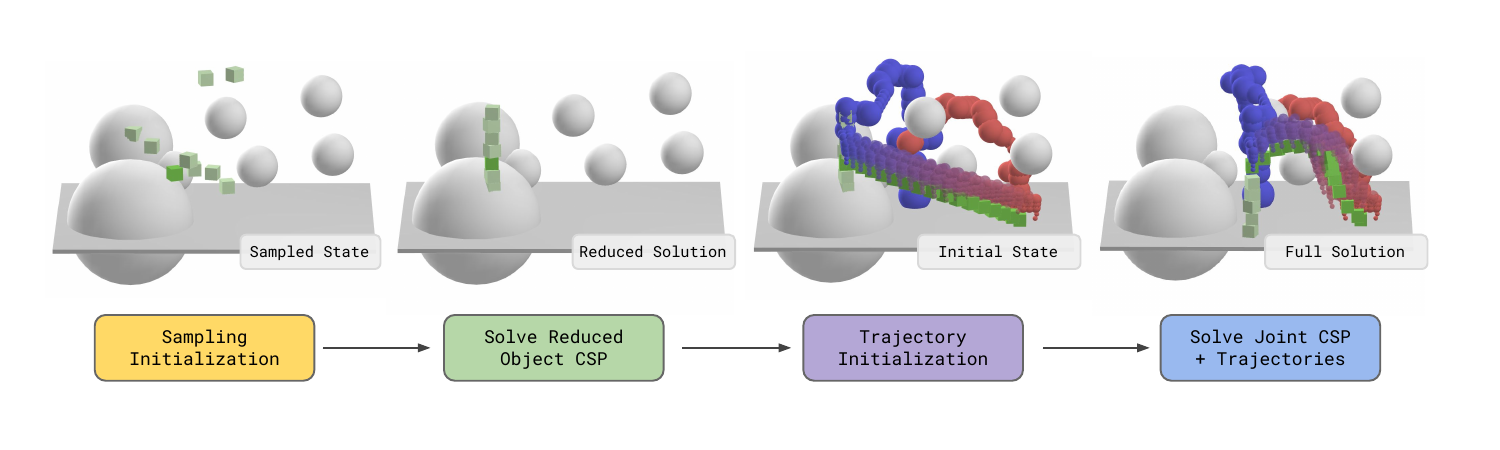}
    \caption{Overview of our approach, Sampling Particle optimization for Sequential Manipulation (SPaSM), applied to our 10-block tower problem in clutter. SPaSM jointly optimizes over the robot configuration and object states for sequential pick and place tasks using large scale particle optimization on the GPU, solving end-to-end placements and motions for this problem in 12 milliseconds.}
    \label{fig:placeholder}
    \vspace{-1em}
\end{figure*}

Robotic manipulation tasks require executing sequences of coordinated motions where each action fundamentally depends on the successful completion of preceding actions.
These dependencies arise from geometric constraints imposed by object grasping configurations, kinematic limitations of robot arms, and obstacles that restrict feasible motion paths.
The coupling between action grounding (i.e., placement of the objects) and continuous motion planning creates a computationally challenging problem where local failures can cascade through the entire action sequence, resulting in failure to find feasible motion.

Recent advances in GPU-accelerated motion planning and manipulation planning such as cuRobo~\cite{sundaralingam2023curobo1} and cuTAMP~\cite{shen2025differentiable} have demonstrated significant computational improvements by parallelizing operations across thousands of concurrent threads.
These methods leverage GPU hardware to accelerate computation.
However, there are still bounding CPU-based coordination steps which incur overhead from CPU-GPU data transfers.
The computational bottleneck has shifted from individual constraint evaluations to the coordination and data movement between processing units.

To address this, we present SPaSM (Sampling Particle Optimization for Sequential Manipulation), a GPU-parallel framework that performs end-to-end optimization of manipulation trajectories entirely on GPU hardware.
We focus on the problem of finding collision-free feasible motion sequences that satisfy a predetermined sequence of manipulation actions.
Given a fixed action skeleton specifying the order of pick-and-place operations, our approach jointly optimizes object placements and robot trajectories to minimize execution cost while satisfying all geometric and kinematic constraints.
The resulting problem structure admits massive parallelization across both candidate solutions and optimization iterations, with zero CPU coordination overhead.
Our method uses a two-stage approach: first optimizing object placement configurations through parallel particle optimization, then lifting successful placements to full trajectory optimization in joint space.
Unlike previous approaches, our second phase performs joint optimization of robot motion and object positions to handle scenarios where motion feasibility significantly constrains placement options.
The entire optimization---including sampling, constraint evaluation, gradient computation, and trajectory refinement---executes as compiled CUDA kernels without CPU intervention, leveraging JAX~\cite{bradbury2018jax} for efficient implementation.
We provide our implementation open source\footnote{\url{https://github.com/CoMMALab/SPaSM}}.

Evaluation on challenging benchmarks demonstrates that SPaSM achieves solution times in \emph{milliseconds} on complex benchmark problems, representing a $4000\times$ speedup compared to cuTAMP (without action skeleton search) while maintaining 100\% success rates. 
Our approach scales effectively and is capable of solving reactive manipulation scenarios where environmental conditions change.
These performance gains enable real-time replanning capabilities and open new possibilities for manipulation tasks requiring rapid adaptation to dynamic environments.

\section{Related Work}

Sequential robot manipulation tasks require coordinated planning across both discrete action sequences and continuous motion trajectories.
In this work we focus on finding feasible geometric motions and action groundings (i.e., grasp poses and placement locations) over a given sequence of actions---we assume this sequence is given.
The MoveIt Task Constructor~\cite{gorner2019moveit} addresses similar problems by working from prespecified action sequences to generate executable plans via sampling potential groundings and resolving with motion planning.
Related sequential manifold planning approaches~\cite{englert2021sampling-based} tackle sequences of actions specified as manifold constraints.
We contrast this assumption against Task and Motion Planning (TAMP) problems, where algorithms must also find the high-level sequence of actions that have corresponding low-level groundings of said actions~\cite{garrett2021integrated}.
However, both classes of problems require solving for satisfying assignments of the grounding variables, constrained by the geometry of the scene and kinematic feasibility.
The fundamental challenge lies in the coupling between discrete action selection and continuous planning; some approaches tackle this with sampling~\cite{garrett2020pddlstream} or optimization~\cite{quintero-pena2023optimal}.
In this work, we assume a valid grounding exists and attempt to find solutions through highly parallel differentiable particle optimization, quickly exploring the potential space of groundings.

Relatedly, trajectory optimization approaches~\cite{schulman2014motion} formulate planning as an optimization problem, which naturally captures the complex constraints imposed over long horizon problems.
Similar to our approach in formulation is that of logic-geometric programming~\cite{toussaint2015logic-geometric,toussaint2018differentiable}, which also optimizes over both poses of objects and the entire trajectory in a three-layer approach. 
Optimization-based approaches for grounding leverage gradient information but face challenges with highly nonlinear constraints, requiring good initialization to avoid local minima.
We avoid this by leveraging parallelism to sample many initial seeds.

There has also been recent progress in receding horizon sequential planning, using model-predictive control frameworks ~\cite{braun2022rhh,toussaint2022sequence,yan2024impact,zhang2024multi}.
In contrast, our approach plans an entire trajectory from start-to-finish.
It is of interest to adapt our method to the receding horizon case.

\subsection{Parallelization for Planning}

Parallel planning has been desirable since the start of the field~\cite{lozano1991parallel} due to the computational intensity of high-dimensional search problems.
Recently, geometric motion planning has been demonstrated to work in the microsecond regime, opening new possibilities for fast and efficient planning~\cite{thomason2024motions}.
Many parts of planning are embarrassingly parallel~\cite{amato1999probabilistic}.
Early GPU implementations focused on specific components such as parallel RRT variants~\cite{bialkowski2011massively} and collision checking acceleration~\cite{pan2012gpu}.
Recent GPU planners (e.g.,~\cite{huang2025prrtc}) parallelize multiple components of the planner to take advantage of all parts of the hardware.
We use JAX~\cite{bradbury2018jax} to automatically compile our approach into an efficient kernel that can be parallelized.

Trajectory optimization can be accelerated effectively on the GPU~\cite{wu2016parallel,pan2019gpu,heinrich2015real,bu2024symmetric}; recent approaches, such as cuRobo~\cite{sundaralingam2023curobo1}, achieve significant speedups for motion planning.
Planning formulated as optimal transport problems~\cite{le2023accelerating} and Global Tensor Motion Planning~\cite{le2025global} exploit tensor operations for massive parallelization.
These methods, and others such as PyRoki, leverage JAX for differentiable robot planning~\cite{kim2025pyroki}.

For sequential manipulation specifically, recent probabilistic approaches like STAMP~\cite{lee2025stamp} apply parallel simulation and Stein variational gradient descent to explore multiple potential action sequences.
cuTAMP~\cite{shen2025differentiable} represents current state-of-the-art GPU acceleration for long-horizon problems, using differentiable collision checking and constraint satisfaction.
However, cuTAMP requires CPU-based coordination to maintain dynamic constraint sets as necessary for solving TAMP problems, creating communication overhead that limits speedups to seconds rather than milliseconds.
The hierarchical decomposition between placement optimization and trajectory planning in cuTAMP also prevents joint optimization of interdependent constraints.
This also prevents joint optimization of the final trajectory.

Our approach differs fundamentally by eliminating CPU-GPU coordination through end-to-end compiled CUDA kernels, jointly optimizing placement and trajectory feasibility rather than hierarchical decomposition, and achieving millisecond performance.
SPaSM also recognizes that motion feasibility can significantly constrain placement options, and thus we integrate optimization across both discrete grounding and continuous trajectory variables.

\section{Methodology}

We address sequential pick-and-place problems where the geometric properties of all objects, their initial positions, and environmental obstacles are known.
This encompasses packing problems (shown in \cref{fig:tetrisprob}), block stacking tasks (\cref{fig:placeholder}), and other manipulation scenarios requiring multiple coordinated pick-and-place actions.
We assume the action skeleton---the predetermined sequence of high-level actions---is provided as input.
The objective is to determine collision-free object placements that satisfy a constraint function $|f(x)| < \epsilon$, which defines the set of valid intermediate and final configurations, while finding feasible collision-free motion of the manipulator and potentially the grasped object.
We formulate collision constraints as penetration distance penalties between object pairs.
Without loss of generality, we represent all objects and the robot as collections of spheres, which enables efficient computation.
For manipulator motion, we optimize trajectory cost in terms of distance in the robot's configuration space.

Solving this problem reduces to a Constraint Satisfaction Problem (CSP)~\cite{garrett2021integrated}, requiring optimization over continuous variables subject to multiple constraints, which generates a highly nonlinear optimization landscape.
For instance, stacking a tower of $N_{\text{obj}}$ blocks requires optimization over approximately $\mathcal{O}(R^d \times N_{\text{timestep}} \times N_{\text{obj}} \times R^{od})$ variables, where $R^d$ represents the robot's degrees of freedom, $R^{od}$ denotes the object state dimensionality, and $N_{\text{timestep}}$ indicates the number of waypoints per robot trajectory.
To address the high dimensional, non-linear nature of the problem, we use a two-stage optimization approach:
\begin{enumerate}
    \item \textbf{Object Variable CSP}: We solve the reduced CSP for object placement states, obtaining a batch of $N$ candidate solutions that provide collision-free feasible object configurations.
    \item \textbf{Trajectory Optimization}: We lift these placement solutions into the full configuration space of the robot, jointly optimizing robot configurations, trajectories, and object placements.
\end{enumerate}
Our approach is inspired by the particle optimization proposed by cuTAMP~\cite{shen2025differentiable}.
However, we do not solve for entire task and motion plans.
We assume the sequence of actions is given.
In contrast, we jointly optimize over entire robot trajectories and object placements, whereas cuTAMP only solves the object variable CSP and uses cuRobo to resolve motion.
Our approach takes advantage of the known problem structure and uses fused computational kernels for better hardware utilization, resulting in over $4000\times$ faster solve times for the object variable CSP.

\begin{algorithm}[b]
\begin{footnotesize}
\caption{GPU-Parallelized Particle Optimization}
\label{alg:gpu_particle_opt}
\begin{algorithmic}[1]
\Require CSP problem $\Pi$, $N$ sampling batch size, $M$ optimization batch size, $K_{\text{lin}}$ linear optimization steps, $K_{\text{quad}}$ quadratic optimization steps, $\eta_{\text{init}}$ initial learning rate

\State $\mathcal{S} \gets \textsc{SampleUniform}(\Pi, N)$ \Comment{Sample initial particles in parallel}
\State $\mathcal{C} \gets \textsc{Cost}(\mathcal{S})$ \Comment{Evaluate costs for all particles}
\State $\mathcal{I} \gets \textsc{ArgSort}(\mathcal{C})[:M]$ \Comment{Select indices of $M$ best particles}
\State $\mathcal{S}_{\text{opt}} \gets \mathcal{S}[\mathcal{I}]$

\State \Comment{\textbf{Linear Cost Descent}}
\For{$k = 1, \dots, K_{\text{lin}}$}
    \State $\eta \gets \eta_{\text{init}} \cdot (1 - k / K_{\text{lin}})$ \Comment{Apply learning rate schedule}
    \State $\mathcal{G} \gets \nabla_{\mathcal{S}_{\text{opt}}} \textsc{Cost}_{\text{linear}}(\mathcal{S}_{\text{opt}})$
    \State $\mathcal{S}_{\text{opt}} \gets \mathcal{S}_{\text{opt}} - \eta \cdot \mathcal{G}$ \Comment{Perform gradient descent step}
\EndFor

\State \Comment{\textbf{Quadratic Cost Refinement}}
\For{$k = 1, \dots, K_{\text{quad}}$}
    \State $\mathcal{G}_{\text{quad}} \gets \nabla_{\mathcal{S}_{\text{opt}}} \textsc{Cost}_{\text{quadratic}}(\mathcal{S}_{\text{opt}})$
    \State $\mathcal{S}_{\text{opt}} \gets \mathcal{S}_{\text{opt}} - \alpha \cdot \mathcal{G}_{\text{quad}}$ \Comment{Refine with small fixed learning rate $\alpha$}
\EndFor
\\
\State $\mathcal{S}_{\text{satisfying}} \gets \textsc{GetSatisfyingParticles}(\mathcal{S}_{\text{opt}})$
\If{$\mathcal{S}_{\text{satisfying}} \neq \emptyset$}
    \State $\mathcal{C}_{\text{final}} \gets \textsc{Cost}(\mathcal{S}_{\text{satisfying}})$ \Comment{Re-evaluate final costs}
    \State $I_{\text{best}} \gets \textsc{ArgSort}(\mathcal{C}_{\text{final}})[:P]$ \Comment{Return the best $P$ solutions}
    \State \Return $\mathcal{S}_{\text{satisfying}}[I_{\text{best}}]$
\EndIf

\State \Return $\textsc{Failure}$
\end{algorithmic}
\end{footnotesize}
\end{algorithm}

\subsection{Object Variable CSP}
\label{sec:pscsp}

We first discuss the reduced CSP involving only object placement states.
This stage determines the final object configurations such that: (i) they maintain collision-free separation from the environment and other objects, and (ii) they remain within the problem-specified valid bounding constraints.

Our approach uses parallelized differentiable optimization over batches of particles.
Although the optimization variables are continuous, the problem structure often admits only zero-dimensional solution manifolds, making the valid solution set effectively discrete.
For example, tightly packing objects within a constrained region typically yields only a finite number of feasible solutions, as demonstrated by the Tetris problem (\cref{fig:tetrisprob}).
The extreme nonlinearity of these constraints presents a fundamental challenge for continuous optimization methods.

Gradient descent methods in high-dimensional spaces exhibit sensitivity to initialization due to prevalent local minima.
Alternative approaches such as cuTAMP~\cite{shen2025differentiable} attempt to mitigate this through initialization strategies including backtracking, early stopping, and conditional sampling, where satisfying states are sampled in subgraphs before completing the remainder in partial order.
These strategies aim to minimize computational effort on unpromising samples while increasing the probability of sampling near-optimal solution states.
However, such methods introduce branching operations that are poorly suited for GPU acceleration and reduce overall efficiency.
Furthermore, the intrinsic problem structure is not conducive to multi-step conditional approaches.
For instance, in packing problems, object states exhibit strong coupling, meaning subgraph caching or conditional initialization provide minimal improvement.
In other words, promising partial states do not guarantee successful completion of the full state.

Our method demonstrates that increasing batch size provides a more effective approach for CSP search than other particle initialization techniques.
This effectiveness stems from the nature of the parallelism afforded by GPUs, which scale efficiently for simple, branch-free operations such as bulk sampling, rather than complex methods which may result in stalls and coupling between processing units.
To improve success rates, we implement a simple rejection sampling procedure: we sample a large number of particles and immediately eliminate high-cost candidates.
This approach scales effectively and generalizes well to challenging problems with highly coupled states, as it makes no assumptions regarding state dependencies.

\subsubsection{Particle Initialization}

We perform uniform sampling of $N$ initial states within the problem domain and compute the initial cost for each particle using the constraint function $f$.
We select the $M$ particles with the lowest initial costs for subsequent placement optimization.
Following optimization, we verify the existence of successful states and repeat the initialization step if none are found.
This corresponds to lines 1--4 in \cref{alg:gpu_particle_opt}.

\begin{figure}
\vspace{1em}
    \centering
    \includegraphics[width=0.9\linewidth]{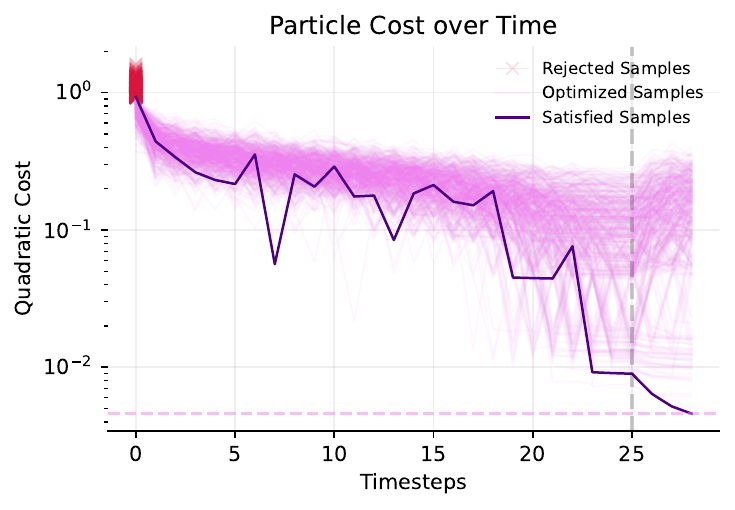}
    \caption{Cost evolution over 30 optimization steps for the 5-block tetris problem, demonstrating the effectiveness of our optimization strategies.
    Satisfying states are frequently (though not universally) identified by low initial costs prior to optimization. 
    Pink trajectories represent samples rejected in the initial step (marked by red particles). 
    Rejected samples rarely converge to valid solutions. 
    At step 25, the transition to quadratic costs reveals states with low costs that cannot actually satisfy the constraints.}
  \label{fig:evolution}
  \vspace{-1em}
\end{figure}

This strategy proves effective because initial state cost serves as a reliable predictor of optimization success.
\cref{fig:evolution} demonstrates the cost evolution across all particles for a representative Tetris problem.
States with higher initial costs exhibit substantially lower probability of successful convergence.
Filtering based on initial cost provides a computationally efficient method for identifying states worthy of optimization investment.
The selection of sampling batch size $N$ and optimization batch size $M$ requires careful balancing for performance.
While larger sample sizes increase the probability that at least one sample will successfully optimize, excessive sampling wastes computational resources since only a single solution is required.
We provide analysis of optimal $N$ and $M$ selections in~\cref{sec:appendix}.

\subsubsection{Particle Optimization}

The selected $M$ particles undergo parallel optimization to determine object placement poses.
During optimization, object poses are clamped to ensure they remain within the problem bounding box.
The highly nonlinear cost landscape induces multiple local minima, and na\"ive cost descent may overlook nearby viable states.
To better exploit the cost landscape structure, we introduce strategies that enhance local neighborhood exploration and improve optimization success probability.
This corresponds to lines 6--23 in \cref{alg:gpu_particle_opt}.

We achieve this through \emph{linear cost formulations} and \emph{learning rate scheduling}.
Initial optimization steps minimize linear formulations of collision costs rather than quadratic alternatives for up to $K_{\text{lin}}$ iterations.
Linear costs enable more aggressive exploration during initial stages by allowing optimization to persist in high-penalty regions if other variables can achieve lower costs.
While quadratic cost growth prevents meaningful navigation of the cost landscape, linear formulations permit escape from local minima, analogous to momentum-based methods.
Subsequently, we transition to quadratic cost optimization for $K_{\text{quad}}$ steps to achieve fine convergence.
As illustrated in \cref{fig:evolution}, introducing quadratic costs at step 25 causes the loss distributions to diverge distinctly into successful and unsuccessful modes.

The second component for landscape exploration uses learning rate scheduling.
The learning rate controls the aggressiveness of exploration away from local minima.
High learning rates enable extensive search for strong attractors (likely solution minima) but prevent convergence to these regions.
The schedule permits high initial exploration uniformly across state variables, followed by increased convergence near the optimization endpoint.
A critical learning rate threshold exists (proportional to the Lipschitz constant and other problem-specific factors) below which the system naturally settles to the nearest local minimum.
We minimize time spent in this regime since it provides limited exploration benefit.
We provide details on our learning rate schedule and other hyperparameters in \cref{sec:appendix}.

\subsection{Trajectory Optimization}

The sampling and optimization procedure from \cref{sec:pscsp} returns feasible states satisfying placement constraints.
In this stage, we use these solutions as initializations and proceed to optimize remaining cost terms, producing feasible robot paths through joint optimization of robot configuration waypoints that form trajectories and object placements.

First, we lift the input object states to joint space representation for each trajectory segment in the motion---this corresponds to a pick or place motion.
We accomplish this by converting object poses to joint space coordinates via sampling inverse kinematics on grasp poses.
In our work, we use the analytic inverse kinematics for the Franka robot proposed by~\citet{he2021analytical}, although any other parallel differentiable inverse kinematics is suitable, e.g.,~\cite{sundaralingam2023curobo1,kim2025pyroki}.
We further initialize our batch of trajectory particles by sampling $K_\text{waypoint}$ intermediate initialization waypoints between each motion segment's start and goal, taking inspiration from~\cite{le2025global} and the retraction state used by cuRobo~\cite{sundaralingam2023curobo1}.
We then initialize trajectories by linearly interpolating between the start, intermediate initialization points, and goal for $K_\text{interp}$ steps for each motion segment.
Finally, we execute gradient descent with learning rate scheduling over the entire trajectory using the gradient of the cost functions (specified in~\cref{sec:appendix}) within an augmented Lagrangian optimization framework~\cite{nocedal2006numerical}.

\section{Experimental Evaluation}

We evaluate SPaSM across a range of sequential manipulation planning problems, including point-to-point motion planning tasks, sequential pick-and-place tasks for packing and tower stacking in clutter, and real-time adaptation scenarios.
These problems vary significantly in difficulty, with particular emphasis on tasks featuring tightly coupled constraints and high nonlinearity.
We compare our approach against state-of-the-art baselines cuTAMP~\cite{shen2025differentiable} and cuRobo~\cite{sundaralingam2023curobo1} (without jerk minimization) across various batch sizes.
All experiments were conducted on an x86-based desktop computer with an AMD Ryzen Threadripper PRO 5965WX 24-Core CPU and an NVIDIA GeForce RTX 4090 GPU.

\subsection{Point-to-Point Motion Planning Problems}

\begin{figure}
\vspace{1em}
    \centering
    \includegraphics[width=0.95\linewidth]{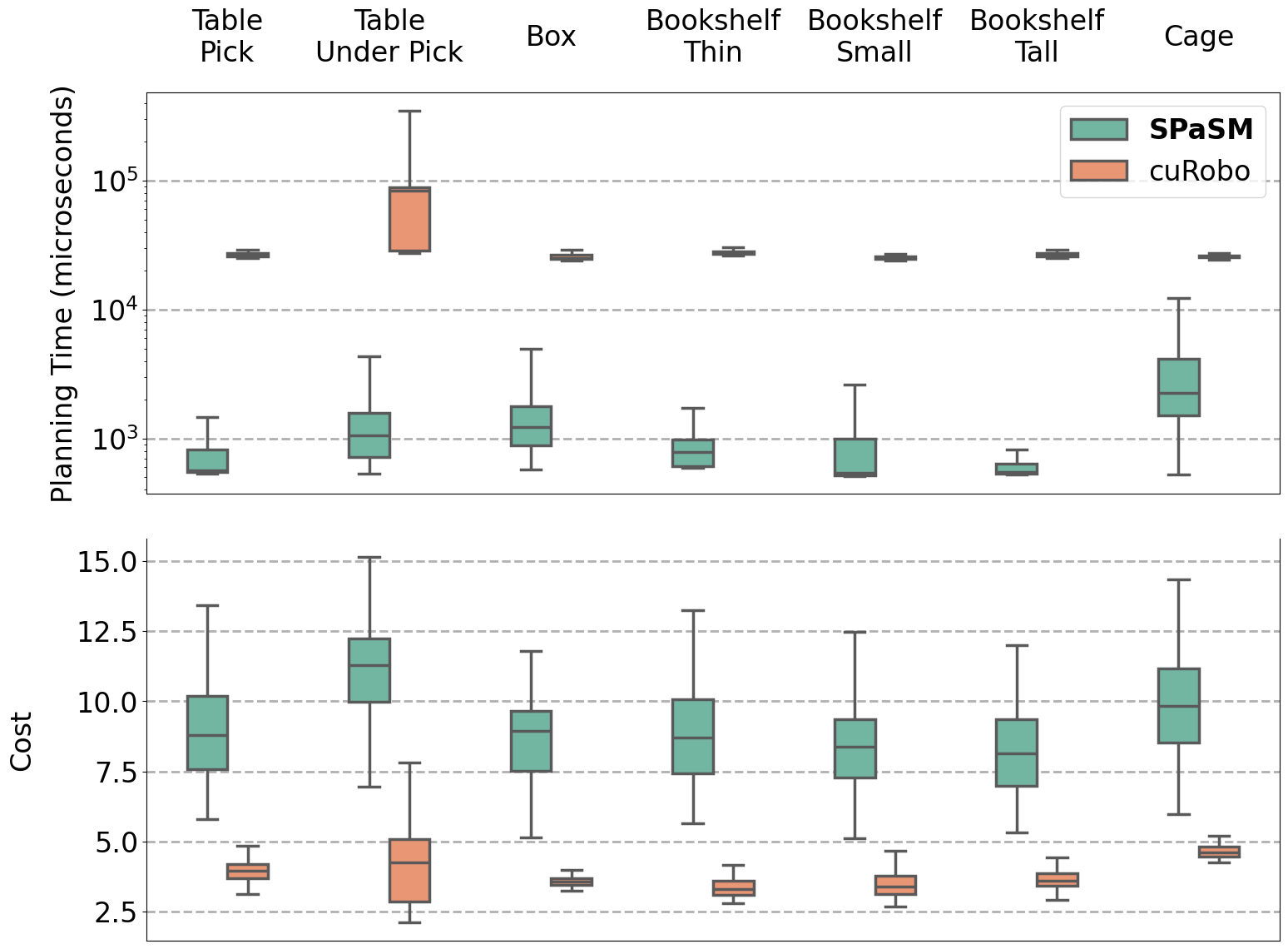}
    \caption{Comparison of SPaSM with cuRobo on the MotionBenchMaker dataset. The success rate of both planners is 100\%. SPaSM is more than an order of magnitude faster, but as expected, found solution paths are longer.}
    \label{fig:mbm}
\end{figure}

\begin{figure}
    \centering
    \includegraphics[width=0.7\linewidth]{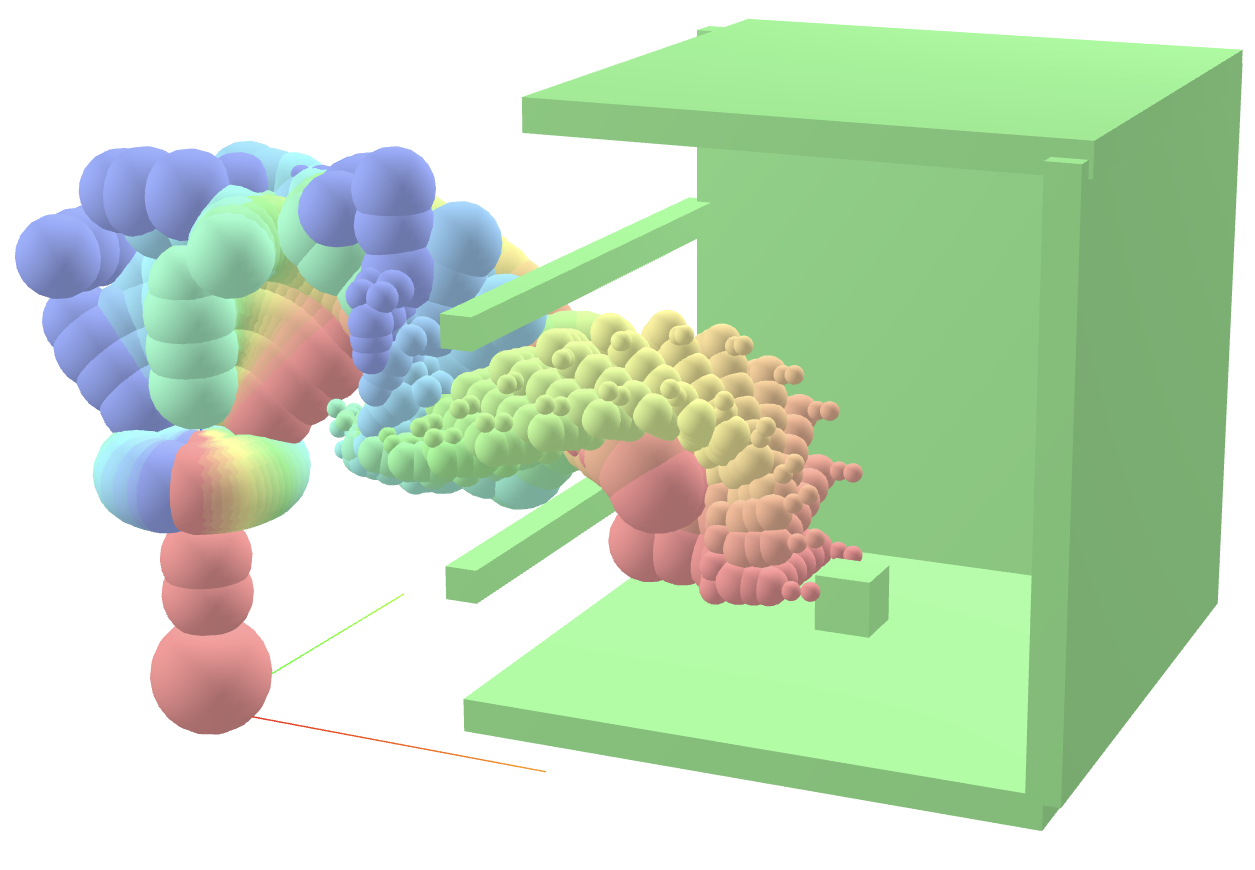}
    \caption{Cutaway with an example trajectory generated for an instance of the ``cage'' problem from MotionBenchMaker~\cite{chamzas2021motionbenchmaker}. SPaSM is able to quickly generate collision free trajectories in complex environments.}
    \label{fig:mbmeg}
    \vspace{-1em}
\end{figure}

\begin{figure}
\vspace{1em}
\centering
  \begin{subfigure}[b]{0.4\columnwidth}
    \frame{\includegraphics[width=\linewidth]{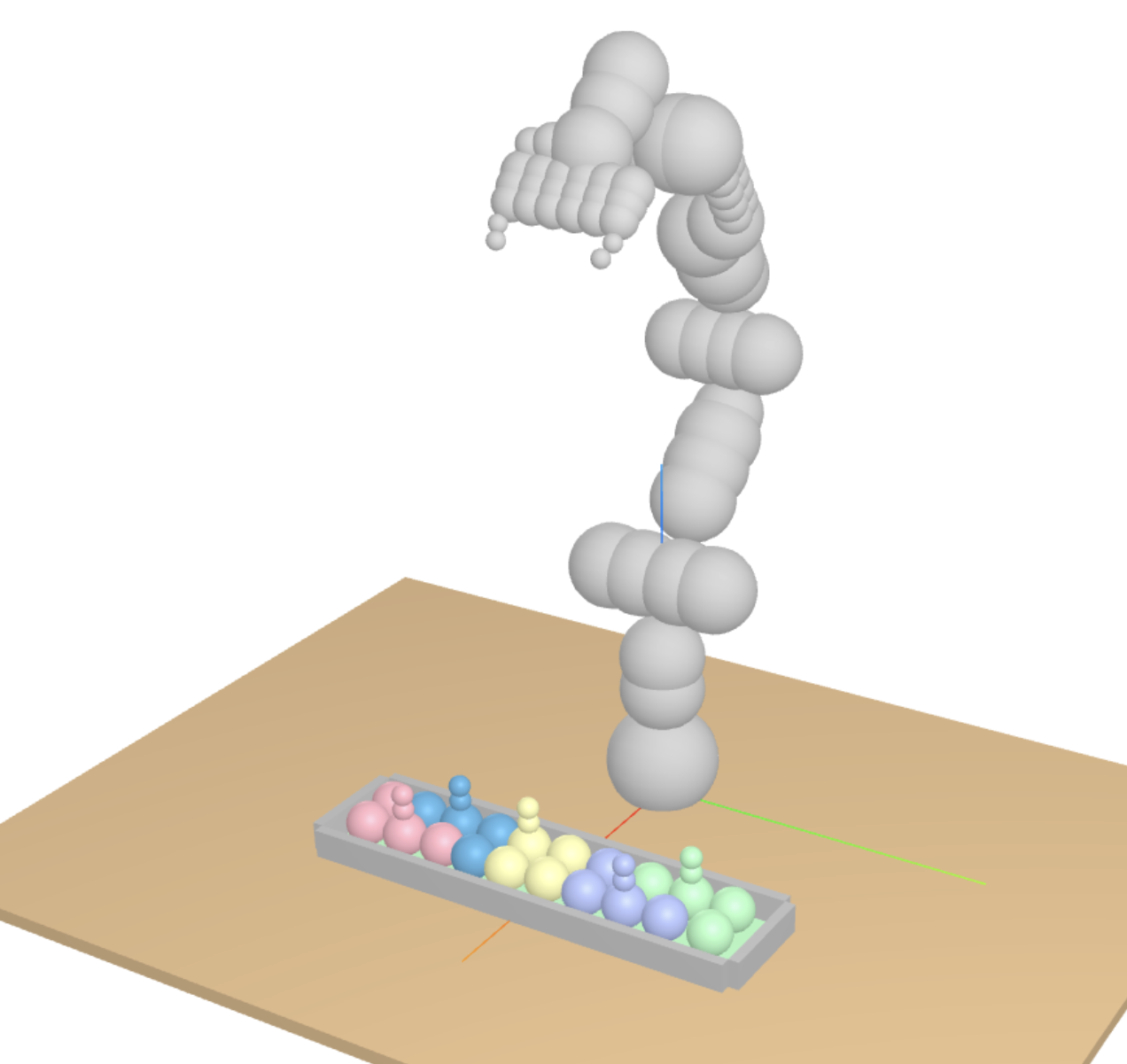}}
        \caption{5 tetris}
    \label{fig:1}
  \end{subfigure}
  \begin{subfigure}[b]{0.4\columnwidth}
    \frame{\includegraphics[width=\linewidth]{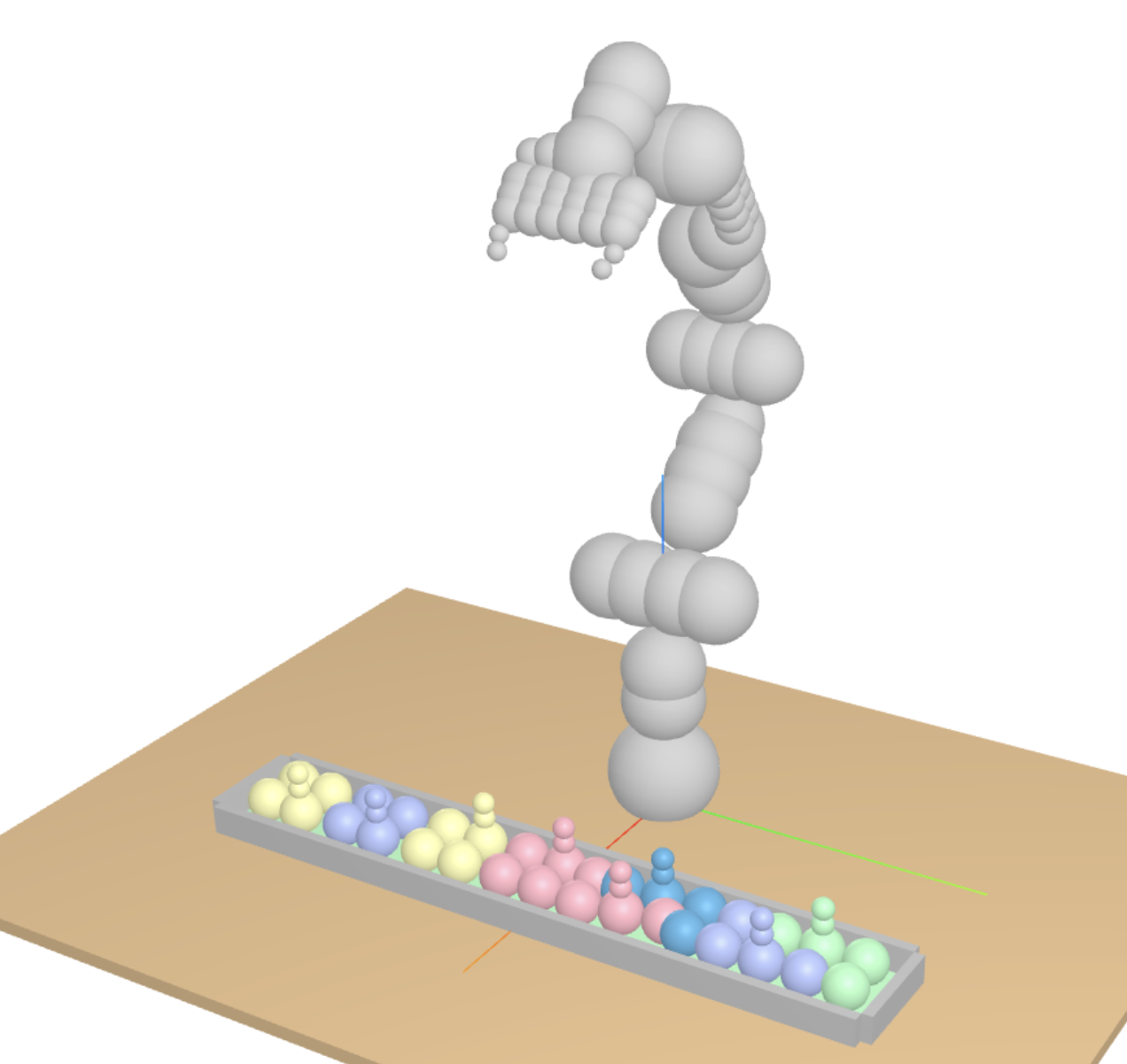}}
        \caption{8-Tetris}
    \label{fig:2}
  \end{subfigure}
    \caption{5-Tetris and 8-Tetris problems. The goal is to pack the blocks into the bounding box. This problem requires }
    \label{fig:tetrisprob}
\end{figure}

The general formulation of SPaSM enables its application to classical motion planning problems by treating them as single-motion trajectory optimization instances.
These problems can be conceptualized as single-sequence-length manipulation tasks, with constraints imposed on feasible configurations by environment geometry.
We evaluate SPaSM with its trajectory optimization component (SPaSM + TrajOpt) on the MotionBenchMaker dataset~\cite{chamzas2021motionbenchmaker} for the 7-DoF Franka Panda.
We compare against cuRobo~\cite{sundaralingam2023curobo1}, a state-of-the-art GPU-accelerated trajectory optimization-based motion planner, measuring both runtime and solution path length.

\cref{fig:mbm} presents the comparative results across different problem instances.
SPaSM demonstrates superior planning time performance, achieving solution times more than an order of magnitude faster than cuRobo while maintaining the same 100\% success rate as cuRobo.
As expected, the solution paths generated by SPaSM are longer than those produced by cuRobo, reflecting a trade-off between optimization speed and path optimality.
These results demonstrate SPaSM's capability to solve difficult planning problems within a unified framework.
Motion planning and manipulation planning can be solved by the same method without losing performance.

\subsection{Pick-and-Place: Tetris Packing}

\begin{table}[t]
\centering

\begin{subtable}[t]{0.5\textwidth} 
\centering
    \begin{tabular}{l|r|cr}
        \toprule
        \textbf{Approach} & $N_b$ & \textbf{Success (\%)} & \textbf{Sol. Time} \\
        \midrule
        \multirow{5}{*}{\parbox{1.35cm}{\scshape SPaSM w/o TrajOpt \\ (tuned)}} 
        & 512 & 100 & 4.23 ± 2.68 ms \\
        & 1024 & 100 & \textbf{4.12 ± 1.94 ms}  \\
        & 2048 & 100 & 4.56 ± 1.71 ms \\
        & 4096 & 100 & 6.37 ± 1.97 ms \\
        & 8192 & 100 & 12.66 ± 2.11 ms \\
        \midrule
        \multirow{5}{*}{\parbox{1.35cm}{\scshape cuTAMP only\\ (tuned)}} 
        & 512 & 100 & 18.98 ± 6.80 s \\
        & 1024 & 100 & 12.14 ± 2.92 s \\
        & 2048 & 100 & \textbf{7.33 ± 0.93 s} \\
        & 4096 & 100 & 9.01 ± 1.10 s \\
        & 8192 & 100 & 11.88 ± 0.98 s \\
        \bottomrule

\end{tabular}
\caption{5-Tetris}
\label{fig:5table}
\end{subtable}

\vspace{1em} 

\begin{subtable}[t]{0.5\textwidth}
\centering
    \begin{tabular}{l|r|cr}
        \toprule
        \textbf{Approach} & $N_b$ & \textbf{Success (\%)} & \textbf{Sol. Time} \\
        \midrule
        \multirow{5}{*}{\parbox{1.35cm}{\scshape SPaSM w/o TrajOpt  \\ (Tuned)}} 
        & 512 & 100 & \textbf{12.68 ± 7.40 ms}  \\
        & 1024 & 100 & 14.15 ± 6.03 ms \\
        & 2048 & 100 & 17.45 ± 5.10 ms \\
        & 4096 & 100 & 36.71 ± 3.70 ms \\
        & 8192 & 100 & 88.97 ± 3.77 ms \\
        \midrule
        \multirow{5}{*}{\parbox{1.35cm}{\scshape cuTAMP only \\ (Tuned)}} 
        & 512 & 80.0 & \textbf{65.48 ± 30.38 s} \\
        & 1024 & 83.33 & 133.69 ± 61.31 s \\
        & 2048 & 100 & 119.91 ± 49.12 s \\
        & 4096 & 100 & 74.58 ± 15.15 s \\
        & 8192 & 100 & 90.06 ± 19.74 s \\
        \bottomrule
        \end{tabular}
            \caption{8-Tetris}
    \label{fig:8table}
\end{subtable}

\caption{Comparison for the 5-Tetris and the 8-Tetris problems in terms of success (coverage) and solution time. Note that for SPaSM times are reported in \textbf{milliseconds} and for cuTAMP in seconds.}
\label{tab:tetrysstage1}
\vspace{-1em}
\end{table}

\begin{table}[t]
\vspace{1.5em}
\centering
    \begin{tabular}{l|r|rr}
        \toprule
        \textbf{Problem} & \textbf{Approach} &  \textbf{Sol. Time} & \textbf{Sol. Length } \\
        \midrule
        \multirow{2}{*}{\parbox{1cm}{\scshape 5-Tetris}} 
        & SPaSM + TrajOpt  & \textbf{10.03 ± 0.95 ms} & \textbf{17.25 ± 0.36}  \\
        & cuTAMP + cuRobo  & 10.55 ± 1.90 s & 28.24 ± 2.47  \\
        \midrule
        \multirow{2}{*}{\parbox{1cm}{\scshape 8-Tetris}} 
        & SPaSM + TrajOpt  & \textbf{19.39 ± 2.06 ms} & \textbf{26.03 ± 0.53}  \\
        & cuTAMP + cuRobo  & 69.42 ± 32.12 s & 48.75 ± 3.15  \\
        \bottomrule

        \end{tabular}
        \caption{Comparison for the full 5-Tetris problem with joint optimization. Both methods have a 100\% success rate. We choose the best batch size based on the results from \cref{fig:tetrisfrontier}.}
\label{tab:tetrysstage2}
\vspace{-1em}
\end{table}

We evaluate SPaSM on a sequential pick-and-place task requiring precise object placement configurations, based on the 5-Tetris benchmark introduced by cuTAMP~\cite{shen2025differentiable}, shown in~\cref{fig:tetrisprob}.
We evaluate scenarios with 5 and 8 Tetris blocks: the planner must determine feasible placements for the blocks that fit within a tight rectangular bounding box (\cref{fig:tetrisprob}).
The blocks are designed such that satisfying packings utilize all available space without gaps, emphasizing that valid solutions constitute effectively a sparsely distributed zero-dimensional subset in the configuration space.
To demonstrate scalability, we developed an 8-block variant extending the original benchmark.
For SPaSM, the cost function combines inter-object penetration penalties with boundedness constraints within the specified region.

\subsubsection{Baseline Configuration}
Since cuTAMP does not perform trajectory optimization within its standard pipeline, we combine it with cuRobo as recommended by the original authors to generate complete motion plans.
We present cuTAMP results both with and without cuRobo trajectory optimization.
Additionally, because our method assumes a given action skeleton, we initialize cuTAMP with a predetermined skeleton known to yield feasible solutions, ensuring fair comparison.
cuTAMP was configured with tuned cost parameters and all visualization disabled for performance.
For results incorporating trajectory optimization, we report runtime and solution cost metrics only for the best-performing batch from stage one, as the batch size for trajectory optimization remains constant regardless of sampling and optimization batch configurations.

\subsubsection{Performance Comparison}
For each problem instance, we conducted 100 trials per method and report the time to first solution with 95\% confidence intervals.
While continued optimization would yield additional solutions, we focus on the more stringent metric of time to first solution, with a maximum limit of 30,000 optimization steps.
Against the cuTAMP and cuRobo baseline, we present results for SPaSM with trajectory optimization and SPaSM without trajectory optimization (stage 1 only).
For SPaSM without trajectory optimization, success is measured solely on placement feasibility, excluding trajectory considerations.

\cref{fig:5table,fig:8table} demonstrate that SPaSM successfully finds solutions in all test cases, while cuTAMP fails on some instances of the 8-block tetris problem.
Altogether, SPaSM with trajectory optimization achieves solution times more than 1,000 times faster than cuTAMP combined with cuRobo.
More significantly, SPaSM without trajectory optimization finds feasible placements nearly \textbf{4,000 times faster} than cuTAMP alone, demonstrating SPaSM's exceptional efficiency in computing feasible object placements.
With trajectory optimization, SPaSM is roughly 1000$\times$ faster than cuTAMP+cuRobo (\cref{tab:tetrysstage2}). In addition, SPaSM generates shorter trajectories due to the joint optimization setup.

\subsubsection{Hyperparameter Ablations}
We conducted ablation studies to determine optimal sampling batch size $N$ and optimization batch size $M$ for SPaSM without trajectory optimization.
We performed hyperparameter search over sampling batch sizes $\in 2^{[9, 19]}$ and optimization batch sizes $\in 2^{[8, 13]}$.
Each parameter combination was evaluated across 100 trials with average solution times recorded.
Our efficient runtime enables such hyperparameter exploration.
\begin{figure}\vspace{1em}
  \begin{subfigure}[b]{0.505\columnwidth}
    \includegraphics[width=\linewidth]{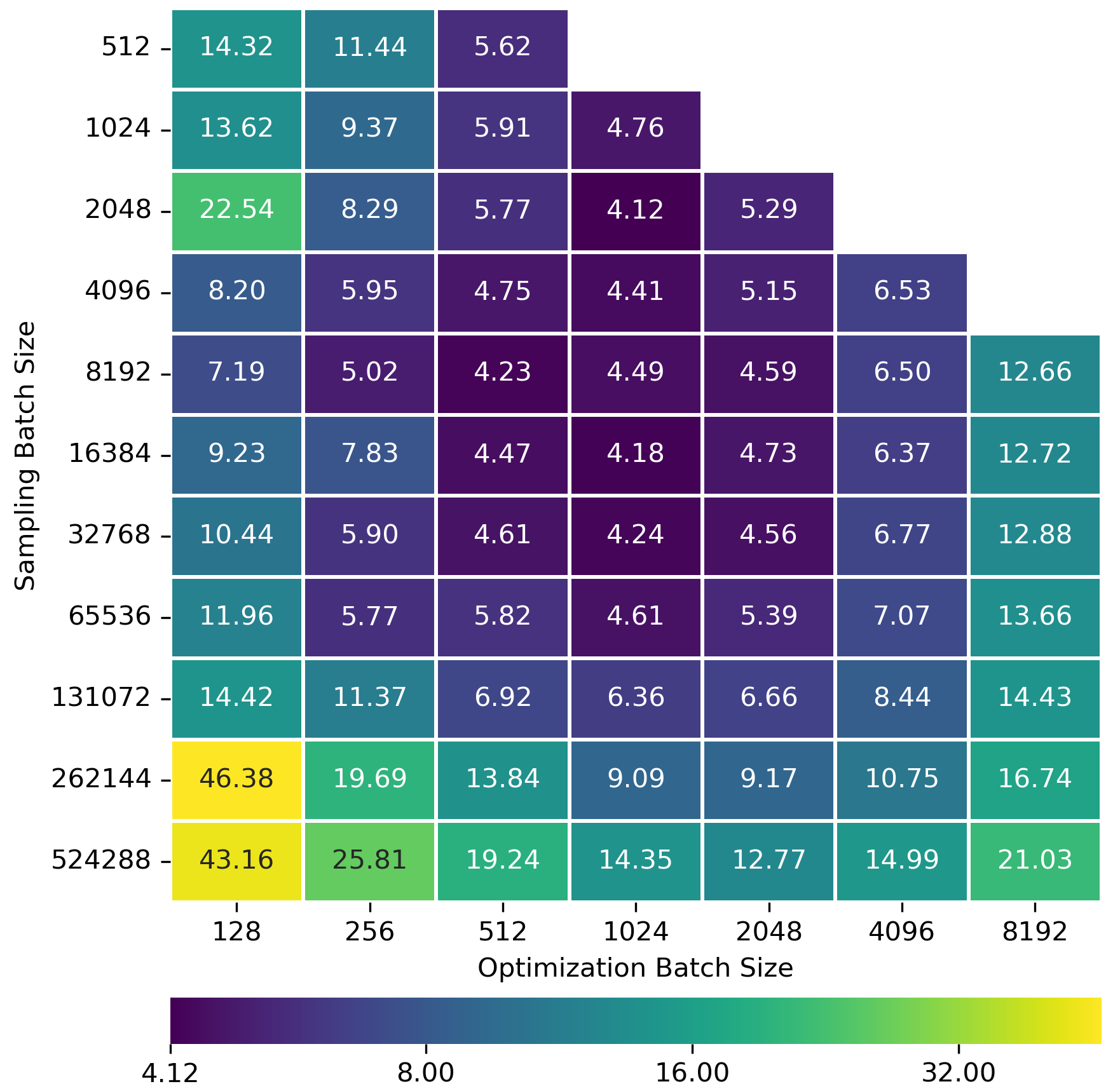}
        \caption{5-Tetris}
    \label{fig:5grid}
  \end{subfigure}
  \hfill 
  \begin{subfigure}[b]{0.485\columnwidth}
    \includegraphics[width=\linewidth]{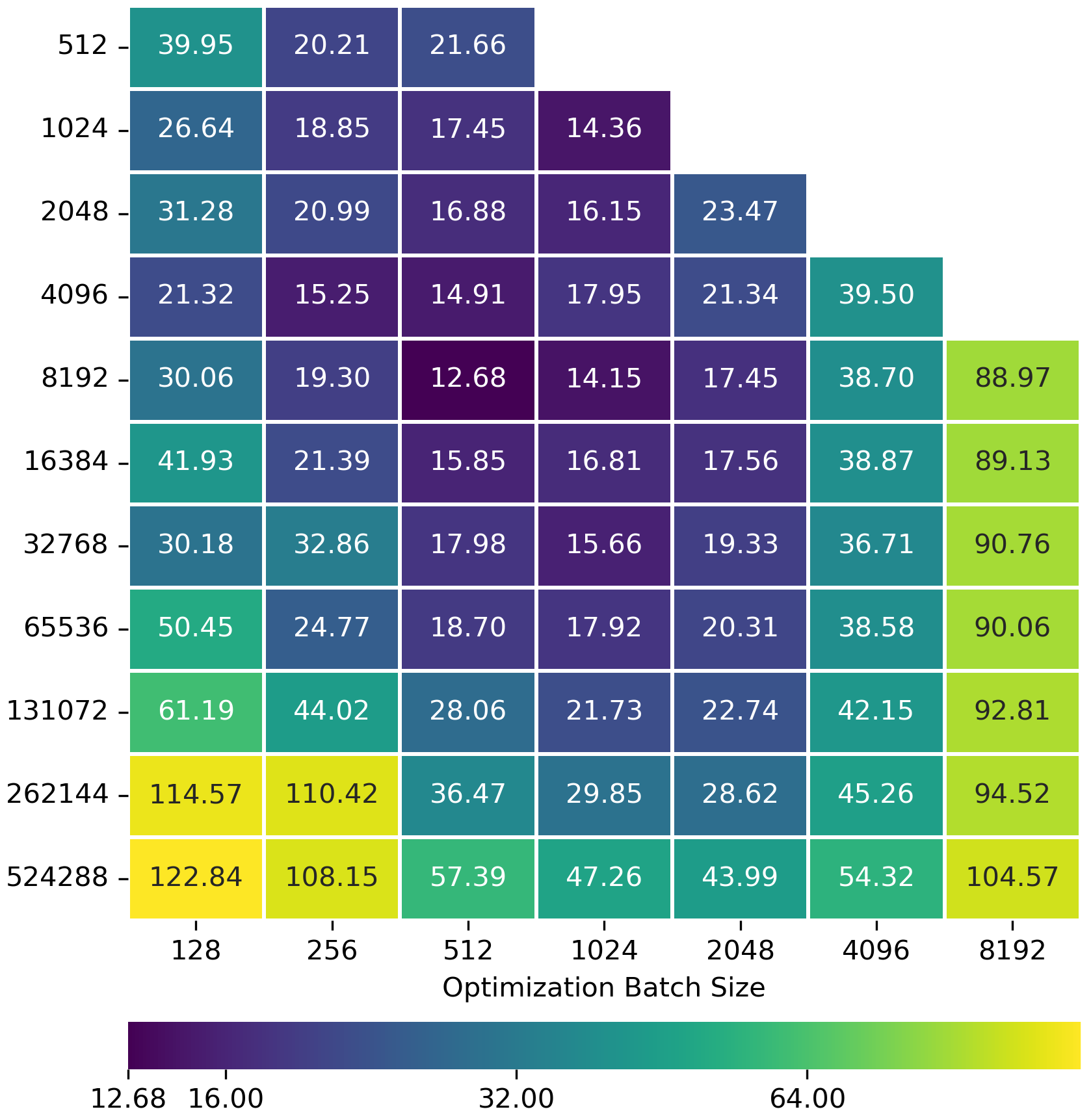}
        \caption{8-Tetris}
    \label{fig:8grid}
  \end{subfigure}
    \caption{Hyperparameter sweep over sampling batch and optimization batch sizes ($N$ and $M$) and average solution times for 5-Tetris and 8-Tetris. There is a balance between high batch sizes (which have higher chance of solution for one solve step, but are computationally heavy) and smaller batch sizes (which have less likelyhood of solution but take less time).}
    \label{fig:tetrisfrontier}
\end{figure}

The analysis revealed unexpected relationships between these parameters (\cref{fig:5grid,fig:8grid}).
Sampling incurs significantly lower computational cost than optimization (1 cost evaluation per batch versus 30 gradient evaluations per batch).
However, since only a single solution is required, excessive sampling wastes computational resources by generating redundant solutions.
The search identified two distinct optimization strategies for minimizing time to first solution: (1) small batch sizes for both parameters, enabling fast iteration with low re-solving penalties, and (2) large batch sizes for both parameters, providing high success probability on initial attempts.
The more challenging 8-Tetris problem performed better with larger batch sizes, likely correlating with the increased sparsity of solution states.
%

\subsection{Pick-and-Place: Tower Stacking}
\begin{figure}
  \begin{subfigure}[b]{0.32\columnwidth}
    \frame{\includegraphics[width=\linewidth]{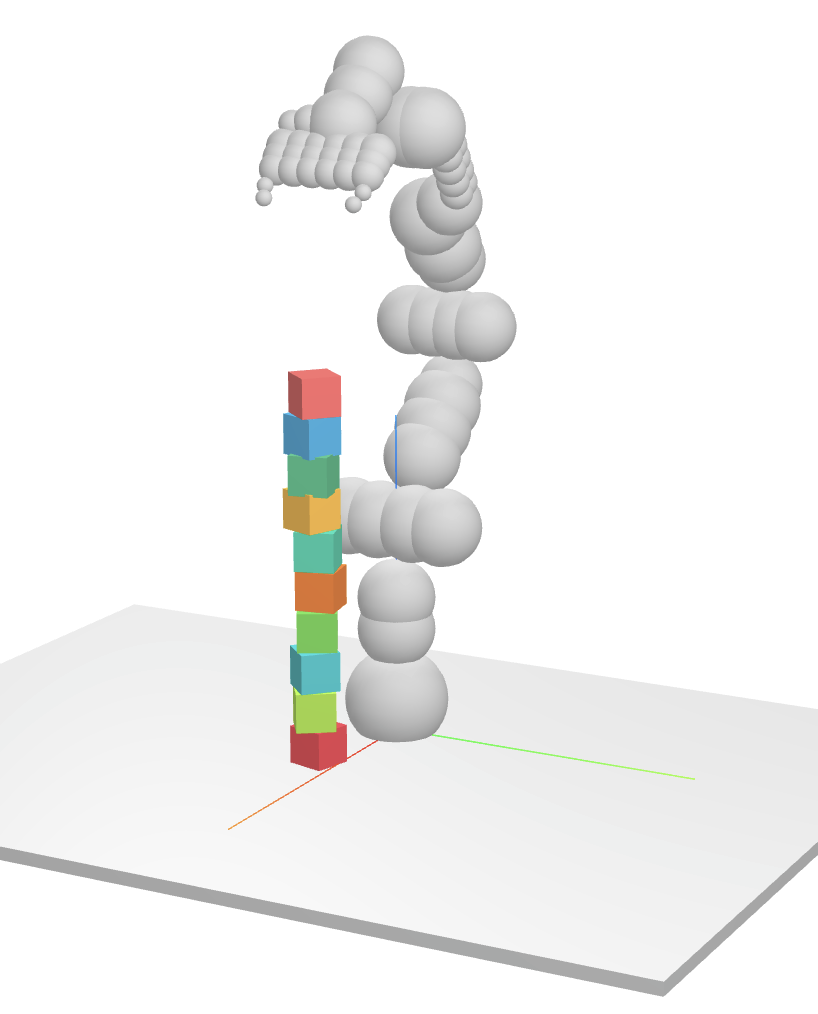}}
        \caption{No Obstacles}
    \label{fig:1}
  \end{subfigure}
  \hfill 
  \begin{subfigure}[b]{0.32\columnwidth}
    \frame{\includegraphics[width=\linewidth]{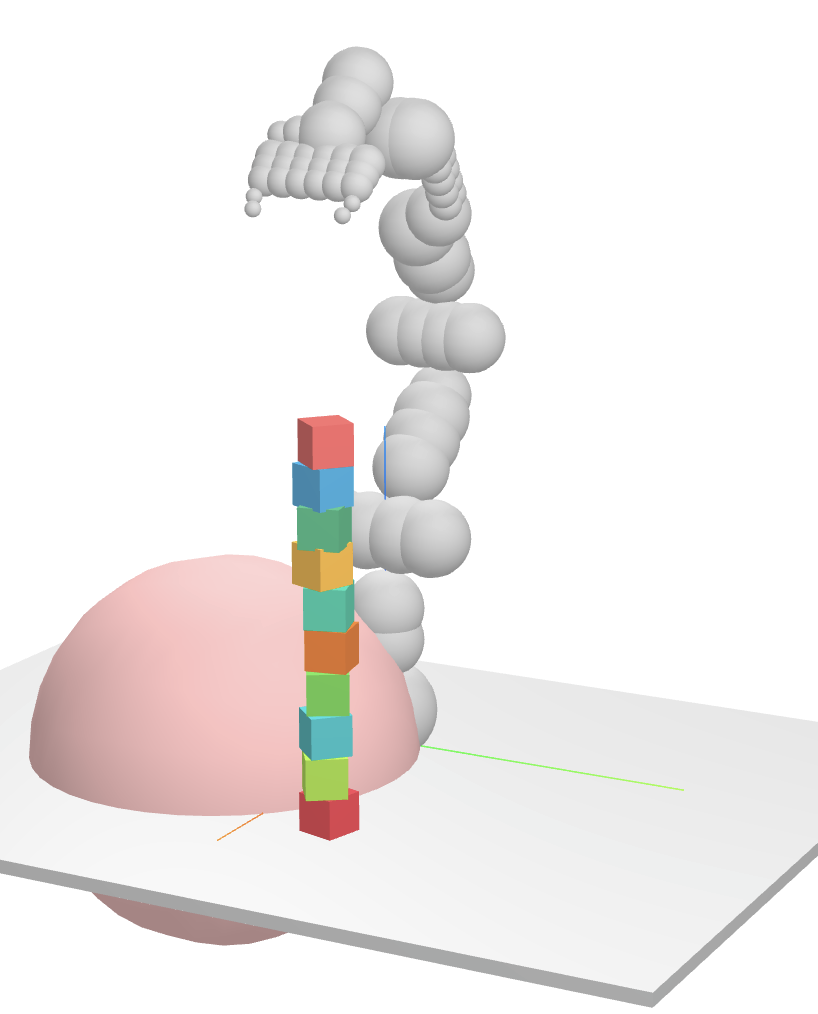}}
        \caption{One Obstacle}
    \label{fig:2}
  \end{subfigure}
  \begin{subfigure}[b]{0.32\columnwidth}
    \frame{\includegraphics[width=\linewidth]{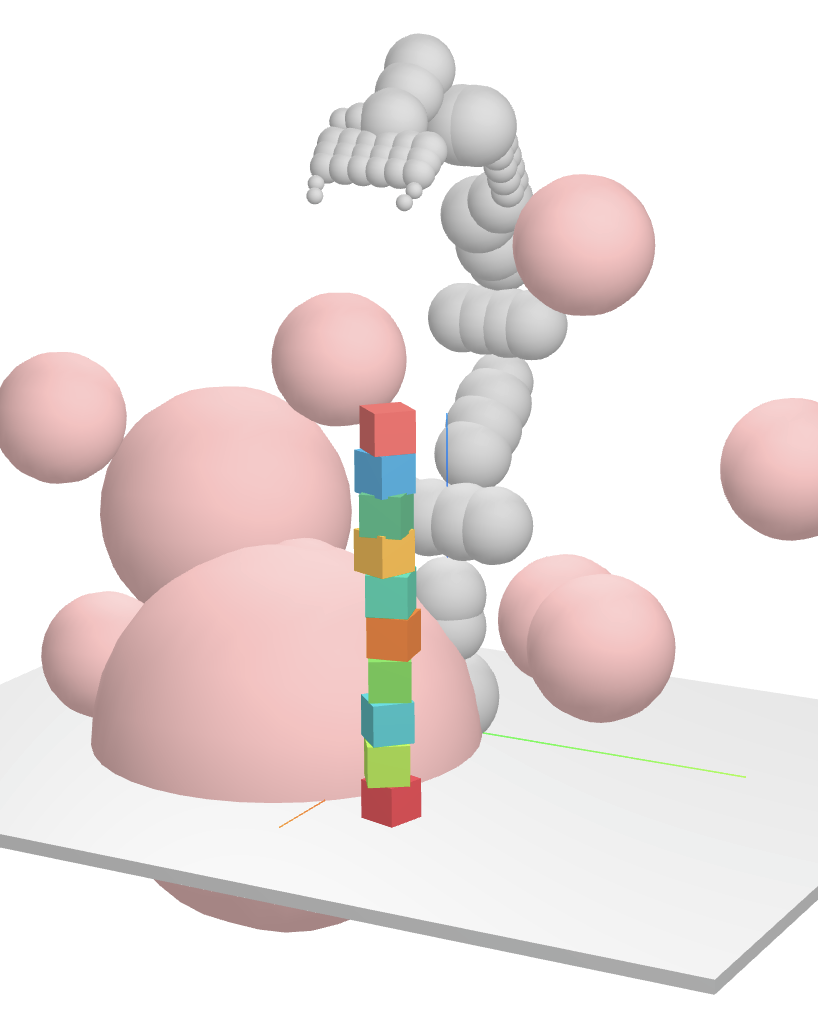}}
        \caption{Ten Obstacles}
    \label{fig:2}
  \end{subfigure}
    \caption{Tower stacking problem. The goal is to stack blocks on the table into a tower while avoiding obstacles. Solving this problem requires joint optimization of placement and trajectory to find feasible end-to-end motions.}
    \label{fig:towerprob}
\vspace{-1em}
\end{figure}

\begin{table}[b]
\vspace{1em}
\centering
    \begin{tabular}{r|rrrr}
        \toprule
        \parbox{0.8cm}{\textbf{Num. \\Obs.}} & 
        \textbf{Sol. Time} &
        \textbf{Sol. Length }  &
        \parbox{1.3cm}{\textbf{Particle \\ Opt. Time}} &
        \parbox{1.3cm}{\textbf{TrajOpt \\ Time}} \\
        \midrule
        0 & 10.80 ± 0.07 & 33.21 ± 0.84 & 0.39 ± 0.01 & 10.42 ± 0.06 \\
        \midrule
        1 & 10.75 ± 0.07 & 32.49 ± 0.66 & 0.38 ± 0.01 & 10.37 ± 0.06 \\
        \midrule
        10 & 12.76 ± 0.06 & 32.99 ± 0.70 & 0.40 ± 0.01 & 12.36 ± 0.05 \\
        \bottomrule

        \end{tabular}

\caption{Ablation studies for the 10-block tower stacking problem with different number of obstacles (\cref{fig:towerprob}). It is interesting to note that SPaSM+TrajOpt scales well even with increasing environment complexity.}
\label{tab:tower}
\end{table}

Combining the computational challenges from point-to-point motion planning and discrete placement optimization, we designed a tower stacking problem involving long-horizon sequential manipulation under environmental constraints.
We address the task of stacking $B=10$ blocks into a physically stable tower configuration.
We evaluate three problem variants: without obstacles,  with a single obstacle, and  with multiple obstacles (\cref{fig:towerprob}).
The obstacles constrain the robot's motion, rendering many potential tower placements infeasible.
Our joint optimization framework simultaneously solves for stable block placement (incorporating center-of-mass constraints and collision-free object positioning) and collision-free arm trajectories, thus enabling solutions in cluttered environments.

For the tower stacking problem, even without obstacles, SPaSM without trajectory optimization consistently fails because interpolated straight-line trajectories inevitably collide with previously placed blocks.
This constraint emphasizes the necessity of trajectory optimization for sequential manipulation tasks.
Following the evaluation done for tetris, we benchmark runtimes and solution lengths across all three problem variants (\cref{tab:tower}).

Our method successfully solves the multiple-obstacle configuration with a 100\% success rate, demonstrating robustness to environmental clutter.
\cref{tab:tower} shows that solution times remain consistent across obstacle configurations, indicating that our approach scales effectively with increasing complexity.
The particle optimization phase requires approximately 0.39 milliseconds regardless of obstacle density, while trajectory optimization time takes between 10 and 13 milliseconds.
This consistency demonstrates SPaSM's scalability and robustness to complex environments and tasks.

\section{Conclusion and Future Work}

We present Sampling Particle optimization for Sequential Manipulation (SPaSM), a fully GPU-parallelized method for sequential pick-and-place planning tasks in clutter that achieves solution times on the order of \emph{milliseconds} on challenging benchmark problems---a $4000\times$ speedup compared to existing approaches~\cite{shen2025differentiable}.
Our end-to-end compilation of constraint evaluation, sampling, and gradient-based optimization into optimized CUDA kernels eliminates CPU-GPU coordination overhead that previously limited accelerated planning approaches.
Our approach also jointly optimizes object placements and robot trajectories in a two-stage approach, demonstrating that the coupling between placements and motion requires integrated optimization to solve in cluttered environments.

The millisecond planning horizon achieved by SPaSM on challenging benchmarks questions assumptions about what could be planned online versus offline---SPaSM transforms sequential manipulation into a real-time planning capability, enabling systems to respond to changes in the environment with replanning frequencies sufficient for dynamic adaptation.
This will enable entirely new applications of complex manipulation, including human-robot collaboration, reactive behaviors, and closed-loop control informed by long-horizon information.
Future research directions include integration with perception systems for reactive planning in the real world, extension to receding horizon formulations that integrate robot dynamics into the problem formulation, and incorporation of learned components to avoid collisions and draw potential placement and grasp samples.
We also wish to investigate a multi-robot sequential planning (e.g., as in~\citet{lai2025roboballet}) extension of SPaSM.

\printbibliography{}

@inproceedings{he2021analytical,
	title        = {Analytical Inverse Kinematics for Franka Emika Panda – a Geometrical Solver for 7-DOF Manipulators with Unconventional Design},
	author       = {He, Yanhao and Liu, Steven},
	year         = 2021,
	booktitle    = {Int. Conf. on Control, Mechatronics and Autom.},
	pages        = {194--199},
	doi          = {10.1109/ICCMA54375.2021.9646185}
}

@article{lee2025stamp,
	title        = {STAMP: Differentiable Task and Motion Planning via Stein Variational Gradient Descent},
	author       = {Lee, Yewon and Li, Andrew Z. and Huang, Philip and Heiden, Eric and Jatavallabhula, Krishna Murthy and Damken, Fabian and Smith, Kevin and Nowrouzezahrai, Derek and Ramos, Fabio and Shkurti, Florian},
	year         = 2025,
	journal      = {IEEE Robot. and Autom. Letters},
	volume       = 10,
	number       = 6,
	pages        = {6007--6014},
	doi          = {10.1109/LRA.2025.3561575}
}

@inproceedings{shen2025differentiable,
	title        = {{Differentiable GPU-Parallelized Task and Motion Planning}},
	author       = {William Shen AND Caelan Reed Garrett AND Nishanth Kumar AND Ankit Goyal AND Tucker Hermans AND Leslie Pack Kaelbling AND Tomás Lozano-Pérez AND Fabio Ramos},
	year         = 2025,
	month        = {June},
	booktitle    = {Robotics: Science and Systems},
	address      = {LosAngeles, CA, USA},
	doi          = {10.15607/RSS.2025.XXI.050}
}

@inproceedings{sundaralingam2023curobo1,
	title        = {Curobo: Parallelized collision-free robot motion generation},
	author       = {Sundaralingam, Balakumar and Hari, Siva Kumar Sastry and Fishman, Adam and Garrett, Caelan and Van Wyk, Karl and Blukis, Valts and Millane, Alexander and Oleynikova, Helen and Handa, Ankur and Ramos, Fabio and others},
	year         = 2023,
	booktitle    = {IEEE Int. Conf. on Robot. and Autom.},
	pages        = {8112--8119}
}

@article{le2025global,
	title        = {Global tensor motion planning},
	author       = {Le, An T and Hansel, Kay and Carvalho, Jo{\~a}o and Watson, Joe and Urain, Julen and Biess, Armin and Chalvatzaki, Georgia and Peters, Jan},
	year         = 2025,
	journal      = {IEEE Robot. and Autom. Letters},
	publisher    = {IEEE}
}

@inproceedings{thomason2024motions,
	title        = {Motions in microseconds via vectorized sampling-based planning},
	author       = {Thomason, Wil and Kingston, Zachary and Kavraki, Lydia E},
	year         = 2024,
	booktitle    = {IEEE Int. Conf. on Robot. and Autom.},
	pages        = {8749--8756}
}

@article{huang2025prrtc,
	title        = {prrtc: Gpu-parallel rrt-connect for fast, consistent, and low-cost motion planning},
	author       = {Huang, Chih H and Jadhav, Pranav and Plancher, Brian and Kingston, Zachary},
	year         = 2025,
	journal      = {arXiv preprint 2503.06757}
}

@article{schulman2014motion,
	title        = {Motion planning with sequential convex optimization and convex collision checking},
	author       = {Schulman, John and Duan, Yan and Ho, Jonathan and Lee, Alex and Awwal, Ibrahim and Bradlow, Henry and Pan, Jia and Patil, Sachin and Goldberg, Ken and Abbeel, Pieter},
	year         = 2014,
	journal      = {The Int. Journal of Robotics Research},
	publisher    = {Sage Publications Sage UK: London, England},
	volume       = 33,
	number       = 9,
	pages        = {1251--1270}
}

@article{garrett2021integrated,
	title        = {Integrated task and motion planning},
	author       = {Garrett, Caelan Reed and Chitnis, Rohan and Holladay, Rachel and Kim, Beomjoon and Silver, Tom and Kaelbling, Leslie Pack and Lozano-P{\'e}rez, Tom{\'a}s},
	year         = 2021,
	journal      = {Annual Review of Control, Robotics, and Autonomous Systems},
	publisher    = {Annual Reviews},
	volume       = 4,
	number       = 1,
	pages        = {265--293}
}

@inproceedings{garrett2020pddlstream,
	title        = {Pddlstream: Integrating symbolic planners and blackbox samplers via optimistic adaptive planning},
	author       = {Garrett, Caelan Reed and Lozano-P{\'e}rez, Tom{\'a}s and Kaelbling, Leslie Pack},
	year         = 2020,
	booktitle    = {Int. Conf. on Automated Planning and Scheduling},
	volume       = 30,
	pages        = {440--448}
}

@article{chamzas2021motionbenchmaker,
	title        = {Motionbenchmaker: A tool to generate and benchmark motion planning datasets},
	author       = {Chamzas, Constantinos and Quintero-Pena, Carlos and Kingston, Zachary and Orthey, Andreas and Rakita, Daniel and Gleicher, Michael and Toussaint, Marc and Kavraki, Lydia E},
	year         = 2021,
	journal      = {IEEE Robot. and Autom. Letters},
	publisher    = {IEEE},
	volume       = 7,
	number       = 2,
	pages        = {882--889}
}

@inproceedings{englert2021sampling-based,
	title        = {{Sampling-Based Motion Planning on Sequenced Manifolds}},
	author       = {Peter Englert AND Isabel M {Rayas Fernández} AND Ragesh Kumar Ramachandran AND Gaurav Sukhatme},
	year         = 2021,
	month        = {July},
	booktitle    = {Robotics: Science and Systems},
	address      = {Virtual},
	doi          = {10.15607/RSS.2021.XVII.039}
}

@inproceedings{quintero-pena2023optimal,
	title        = {Optimal grasps and placements for task and motion planning in clutter},
	author       = {Quintero-Pena, Carlos and Kingston, Zachary and Pan, Tianyang and Shome, Rahul and Kyrillidis, Anastasios and Kavraki, Lydia E},
	year         = 2023,
	booktitle    = {IEEE Int. Conf. on Robot. and Autom.},
	publisher    = {IEEE}
}

@inproceedings{toussaint2015logic-geometric,
	title        = {Logic-Geometric Programming: An Optimization-Based Approach to Combined Task and Motion Planning.},
	author       = {Toussaint, Marc},
	year         = 2015,
	booktitle    = {IJCAI},
	pages        = {1930--1936}
}

@inproceedings{toussaint2018differentiable,
	title        = {Differentiable Physics and Stable Modes for Tool-Use and Manipulation Planning},
	author       = {Marc Toussaint AND Kelsey Allen AND Kevin Smith AND Joshua Tenenbaum},
	year         = 2018,
	month        = {June},
	booktitle    = {Robotics: Science and Systems},
	address      = {Pittsburgh, Pennsylvania},
	doi          = {10.15607/RSS.2018.XIV.044}
}

@software{bradbury2018jax,
	title        = {{JAX}: composable transformations of {P}ython+{N}um{P}y programs},
	author       = {James Bradbury and Roy Frostig and Peter Hawkins and Matthew James Johnson and Chris Leary and Dougal Maclaurin and George Necula and Adam Paszke and Jake Vander{P}las and Skye Wanderman-{M}ilne and Qiao Zhang},
	year         = 2018,
	url          = {http://github.com/jax-ml/jax},
	version      = {0.3.13}
}

@inproceedings{lozano1991parallel,
	title        = {Parallel robot motion planning.},
	author       = {Lozano-P{\'e}rez, Tom{\'a}s and O'Donnell, Patrick A},
	year         = 1991,
	booktitle    = {IEEE Int. Conf. on Robot. and Autom.},
	pages        = {1000--1007}
}

@article{pan2012gpu,
	title        = {GPU-based parallel collision detection for fast motion planning},
	author       = {Pan, Jia and Manocha, Dinesh},
	year         = 2012,
	journal      = {The Int. Journal of Robotics Research},
	publisher    = {Sage Publications Sage UK: London, England},
	volume       = 31,
	number       = 2,
	pages        = {187--200}
}

@article{kim2025pyroki,
	title        = {PyRoki: A Modular Toolkit for Robot Kinematic Optimization},
	author       = {Kim, Chung Min and Yi, Brent and Choi, Hongsuk and Ma, Yi and Goldberg, Ken and Kanazawa, Angjoo},
	year         = 2025,
	journal      = {arXiv preprint 2505.03728}
}

@article{le2023accelerating,
	title        = {Accelerating motion planning via optimal transport},
	author       = {Le, An T and Chalvatzaki, Georgia and Biess, Armin and Peters, Jan R},
	year         = 2023,
	journal      = {Advances in Neural Information Processing Systems},
	volume       = 36,
	pages        = {78453--78482}
}

@inproceedings{amato1999probabilistic,
	title        = {Probabilistic roadmap methods are embarrassingly parallel},
	author       = {Amato, Nancy M and Dale, Lucia K},
	year         = 1999,
	booktitle    = {IEEE Int. Conf. on Robot. and Autom.},
	volume       = 1,
	pages        = {688--694}
}

@inproceedings{gorner2019moveit,
	title        = {Moveit! task constructor for task-level motion planning},
	author       = {G{\"o}rner, Michael and Haschke, Robert and Ritter, Helge and Zhang, Jianwei},
	year         = 2019,
	booktitle    = {Int. Conf. on Robot. and Autom.},
	pages        = {190--196}
}

@inproceedings{bialkowski2011massively,
	title        = {Massively parallelizing the RRT and the RRT},
	author       = {Bialkowski, Joshua and Karaman, Sertac and Frazzoli, Emilio},
	year         = 2011,
	booktitle    = {IEEE/RSJ Int. Conf. on Intelligent Robots and Systems},
	pages        = {3513--3518}
}

@article{zhang2024multi,
	title        = {Multi-modal mppi and active inference for reactive task and motion planning},
	author       = {Zhang, Yuezhe and Pezzato, Corrado and Trevisan, Elia and Salmi, Chadi and Corbato, Carlos Hern{\'a}ndez and Alonso-Mora, Javier},
	year         = 2024,
	journal      = {IEEE Robot. and Autom. Letters},
	publisher    = {IEEE}
}

@article{yan2024impact,
	title        = {Impact-aware bimanual catching of large-momentum objects},
	author       = {Yan, Lei and Stouraitis, Theodoros and Moura, Joao and Xu, Wenfu and Gienger, Michael and Vijayakumar, Sethu},
	year         = 2024,
	journal      = {IEEE Transactions on Robotics},
	publisher    = {IEEE},
	volume       = 40,
	pages        = {2543--2563}
}

@inproceedings{toussaint2022sequence,
	title        = {Sequence-of-constraints MPC: Reactive timing-optimal control of sequential manipulation},
	author       = {Toussaint, Marc and Harris, Jason and Ha, Jung-Su and Driess, Danny and H{\"o}nig, Wolfgang},
	year         = 2022,
	booktitle    = {IEEE/RSJ Int. Conf. on Intelligent Robots and Systems},
	pages        = {13753--13760}
}

@inproceedings{braun2022rhh,
	title        = {Rhh-lgp: Receding horizon and heuristics-based logic-geometric programming for task and motion planning},
	author       = {Braun, Cornelius V and Ortiz-Haro, Joaquim and Toussaint, Marc and Oguz, Ozgur S},
	year         = 2022,
	booktitle    = {IEEE/RSJ Int. Conf. on Intelligent Robots and Systems},
	pages        = {13761--13768}
}

@book{nocedal2006numerical,
	title        = {Numerical optimization},
	author       = {Nocedal, Jorge and Wright, Stephen J},
	year         = 2006,
	publisher    = {Springer}
}

@article{lai2025roboballet,
	title        = {RoboBallet: Planning for multirobot reaching with graph neural networks and reinforcement learning},
	author       = {Lai, Matthew and Go, Keegan and Li, Zhibin and Kr{\"o}ger, Torsten and Schaal, Stefan and Allen, Kelsey and Scholz, Jonathan},
	year         = 2025,
	journal      = {Science Robotics},
	publisher    = {American Association for the Advancement of Science},
	volume       = 10,
	number       = 106,
	pages        = {eads1204}
}

@inproceedings{heinrich2015real,
	title        = {Real-time trajectory optimization under motion uncertainty using a GPU},
	author       = {Heinrich, Steffen and Zoufahl, Andre and Rojas, Raul},
	year         = 2015,
	booktitle    = {IEEE/RSJ Int. Conf. on Intelligent Robots and Systems},
	pages        = {3572--3577}
}

@inproceedings{pan2019gpu,
	title        = {Gpu-based contact-aware trajectory optimization using a smooth force model},
	author       = {Pan, Zherong and Ren, Bo and Manocha, Dinesh},
	year         = 2019,
	booktitle    = {ACM SIGGRAPH/Eurographics Symposium on Computer Animation},
	pages        = {1--12}
}

@article{wu2016parallel,
	title        = {Parallel particle swarm optimization on a graphics processing unit with application to trajectory optimization},
	author       = {Wu, Q and Xiong, F and Wang, F and Xiong, Y},
	year         = 2016,
	journal      = {Engineering Optimization},
	publisher    = {Taylor \& Francis},
	volume       = 48,
	number       = 10,
	pages        = {1679--1692}
}

@inproceedings{bu2024symmetric,
	title        = {Symmetric stair preconditioning of linear systems for parallel trajectory optimization},
	author       = {Bu, Xueyi and Plancher, Brian},
	year         = 2024,
	booktitle    = {IEEE Int. Conf. on Robot. and Autom.},
	pages        = {9779--9786}
}

\section{Appendix}
\label{sec:appendix}

\subsection{Tetris Problem}
\label{subsec:tetris_problem}

The placement cost function is defined as:
\begin{equation}
C_{\text{placement}} = \sum_{i \in B} \sum_{j > i \in B} P_{(i, j)} + \sum_{i \in B} \sum_{j \in W} P_{(i, j)} + |z - z^*|
\end{equation}

where $P_{(i,j)}$ represents the penetration depth between objects $i$ and $j$.
The cost includes penetration penalties between block pairs, block-wall pairs, and the distance from the desired z-axis position.
$B$ denotes the set of blocks, $W$ denotes the set of walls, and $z^*$ is the target z-coordinate.

\subsection{Tower Stacking Problem}
\label{subsec:tower_stacking}

The initial placement of blocks is determined by minimizing a cost function $C_{\text{placement}}$ that evaluates the quality of a tower configuration $\{\mathbf{p}_i\}_{i \in B}$.
Here, $\mathbf{p}_i$ represents the pose of block $i$.
The cost function is a weighted sum of penalties for instability, incorrect block height, and collisions:
\begin{equation}
C_{\text{placement}} = w_{\text{stable}}C_{\text{stable}} + w_{\text{height}}C_{\text{height}} + w_{\text{coll}}C_{\text{coll}}
\end{equation}

\noindent\textbf{Stability Cost ($C_{\text{stable}}$):}
Penalizes configurations where the center of mass (CoM) of blocks stacked above block $i$ projects outside its supporting footprint.
\begin{equation}
C_{\text{stable}} = \sum_{i=1}^{B-1} \text{Dist}(\text{CoM}(\{\mathbf{p}_j\}_{j>i}), \text{Footprint}(\mathbf{p}_i))
\end{equation}

\noindent\textbf{Height Cost ($C_{\text{height}}$):}
Enforces that each block $i$ is placed at a height corresponding to its index in the stack.
\begin{equation}
C_{\text{height}} = \sum_{i=1}^{B} (z(\mathbf{p}_i) - i \cdot h_{\text{block}})^2
\end{equation}

\noindent\textbf{Collision Cost ($C_{\text{coll}}$):}
Penalizes penetration between block pairs and between blocks and environmental obstacles $\mathcal{O}$.
\begin{equation}
C_{\text{coll}} = \sum_{i, j > i \in B} \text{Pen}(\mathcal{B}_i, \mathcal{B}_j) + \sum_{i \in B, o \in \mathcal{O}} \text{Pen}(\mathcal{B}_i, o)
\end{equation}

In these equations, $z(\mathbf{p}_i)$ is the vertical position of block $i$, $h_{\text{block}}$ is the height of a single block, and $\text{Pen}(\cdot, \cdot)$ is a penetration depth function.
The optimization is performed over sequences of joint configurations $\mathbf{q}_{b,t} \in \mathbb{R}^7$ for each of the $B$ blocks at each timestep $t \in \{0, \ldots, T\}$:
\begin{align*}
C_{\text{trajopt}} &= C_{\text{placement}}(\{\mathbf{p}_{b,T}\}_{\forall b}) + w_{\text{start}}C_{\text{start}} \\&+ w_{\text{arm}}C_{\text{arm}} + w_{\text{block}}C_{\text{block}}
\end{align*}

\noindent\textbf{Placement Cost ($C_{\text{placement}}$):}
Evaluates the stability of the final tower configuration.
Final block poses $\{\mathbf{p}_{b,T}\}_{\forall b}$ are determined by forward kinematics at the final timestep: $\text{fk}(\mathbf{q}_{b,T})$.

\noindent\textbf{Initial Pose Alignment Cost ($C_{\text{start}}$):}
Penalizes misalignment between the arm's end-effector and the block's initial pose at $t=0$ to ensure smooth grasping.
The cost measures squared error in position and roll/pitch orientation:
\begin{align}
C_{\text{start}} = \sum_{b=1}^{B} &\left\| \text{pos}(\text{fk}(\mathbf{q}_{b,0})) - \text{pos}(\mathbf{p}_{b,\text{initial}}) \right\|^2_2 + \\
&\left\| \text{rot}_{\text{RP}}(\text{fk}(\mathbf{q}_{b,0})) - \text{rot}_{\text{RP}}(\mathbf{p}_{b,\text{initial}}) \right\|^2_2
\end{align}

\noindent\textbf{Arm-Obstacle Collision Cost ($C_{\text{arm}}$):}
Penalizes penetration depth between robot arm geometry $\mathcal{A}(\mathbf{q}_{b,t})$ and workspace obstacles $o \in \mathcal{O}$ across the entire trajectory:
\begin{equation}
C_{\text{arm}} = \sum_{b=1}^{B} \sum_{t=0}^{T} \sum_{o \in \mathcal{O}} \text{Pen}(\mathcal{A}(\mathbf{q}_{b,t}), o)
\end{equation}

\noindent\textbf{Held Block-Obstacle Collision Cost ($C_{\text{block}}$):}
Penalizes penetration depth between held block geometry $\mathcal{G}(\text{fk}(\mathbf{q}_{b,t}))$ and obstacles $o \in \mathcal{O}$ during transport phase ($t \in [1, T-1]$):
\begin{equation}
C_{\text{block}} = \sum_{b=1}^{B} \sum_{t=1}^{T-1} \sum_{o \in \mathcal{O}} \text{Pen}(\mathcal{G}(\text{fk}(\mathbf{q}_{b,t})), o)
\end{equation}

The scalar weights $w_{\text{start}}$, $w_{\text{arm}}$, and $w_{\text{block}}$ balance the contribution of each term.
Minimizing $C_{\text{trajopt}}$ yields joint-space trajectories that achieve stable tower configurations while respecting physical and environmental constraints.

\end{document}